# Hardware In The Loop Simulator in UAV Rapid Development Life Cycle

Widyawardana Adiprawita*, Adang Suwandi Ahmad [=] and Jaka Semibiring[+]

*School of Electric Engineering and Informatics
Institut Teknologi Bandung, Bandung, Id.
e-mail: wadiprwita@stei.itb.ac.id

[=] School of Electric Engineering and Informatics
Institut Teknologi Bandung, Bandung, Id.
e-mail: asa@isrg.itb.ac.id

[+] School of Electric Engineering and Informatics
Institut Teknologi Bandung, Bandung, Id.
e-mail: jaka@ itb.ac.id

**Abstract**

Field trial is very critical and high risk in autonomous UAV development life cycle. Hardware in the loop (HIL) simulation is a computer simulation that has the ability to simulate UAV flight characteristic, sensor modeling and actuator modeling while communicating in real time with the UAV autopilot hardware. HIL simulation can be used to test the UAV autopilot hardware reliability, test the closed loop performance of the overall system and tuning the control parameter. By rigorous testing in the HIL simulator, the risk in the field trial can be minimized.

## 1   Introduction

Unmanned Aerial Vehicle (UAV) has proved to be a very valuable asset in military and commercial application. These applications include reconnaissance, surveillance, search and rescue, remote sensing (nuclear, biological, chemical), traffic monitoring, natural disaster damage assessment, etc. UAV can accomplish its mission without risking the pilot/operator and usually with lower operational cost compared to manned aircraft.

From science and academic point of view, UAV research and development has the potentials of integrating several science and engineering disciplines such as Embedded System, Real Time System, Sensor Fusion and Integration, Power System (for on board power generator), Image Processing (for obstacle avoidance), Information System (for ground control station), Telecommunication (for telemetry), Networking (for information dissemination to remote audience from ground control station), Control System (for automatic airframe stabilization and waypoint navigation), Multi Agent System (for multiple UAVs coordination), etc.

Based on those UAV's potentials, several departments at ITB have been involved in UAV research and development. Specific to School of Electric Engineering and Informatics the objective is to develop a fully autonomous UAV control system prototype which is low cost (using commercially available components), simple to operate, modular, and reliable.

## 2   UAV Development Life Cycle

One of the main component of UAV is its autopilot. The autopilot consists of high level navigation algorithm and low level control law. The common practice of developing control law is based on mathematic model of the airframe (state space or transfer function). This mathematic model is quite difficult to obtain. There are two main approaches of obtaining this mathematic model. First approach is by using first-principle modeling. This first-principle approach involves derivation of airframe's equation of motions from the ground up, using fundamental laws of mechanics and aerodynamics. This requires considerable knowledge and experience with all the phenomena involved in small scale aircraft flight. Unfortunately this approach does not guarantee a highly accurate mathematic models, unless performed with extreme care. The second approach is system identification modeling based on data recorded from real world flight test. This approach proved to be very robust. One of the tools to conduct this approach is CIFER. CIFER is based on frequency domain analysis of the flight test data, so it can give high mathematic model with broad bandwidth accuracy. Because of this approach CIFER need frequency-sweep input to be conducted in flight test. For unmanned aircraft flight test purposes, this frequency-sweep input is not easy to obtain. Beside that flight test is always a high risk procedure.

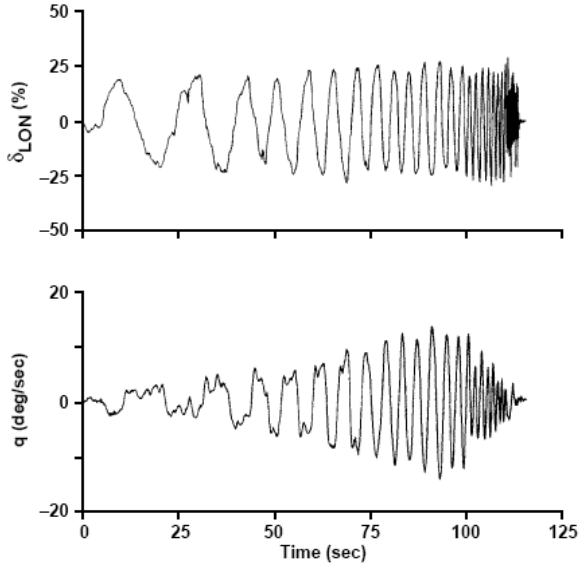

**Figure 1:** Typical input output pair needed for frequency domain system identification in CIFER

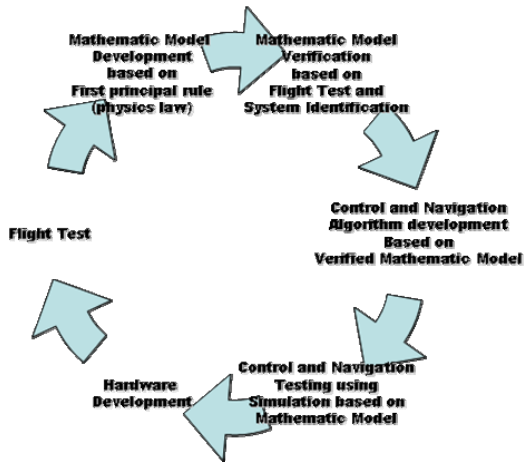

**Figure 2:** Common practice in UAV's control development based on mathematic model

The UAV's control development based on mathematic model proved to be complex and long term effort.

Actually there are many mathematic model for small scale aircraft available. But the form of this mathematic model is not in state space of transfer function formulation. They are available as flight simulation software package. Among them are Microsoft Flight Simulator and X-Plane. The software company have put much effort to make the simulation is as accurate as possible. And usually some the flight simulation software also give some method to access the internal simulation data, so we can interact with the mathematic model implemented inside it.

So, in this research, a rapid methodology will be proposed to develop control system of UAV based on the availability of mathematic model embedded inside commercially available flight simulation software.

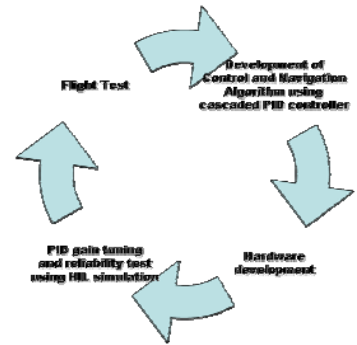

**Figure 3:** Proposed methodology for Rapid UAV development

This proposed rapid development life cycle relies on hardware in the loop simulation. The hardware in the loop simulation will be developed on top of commercially available flight simulation.

## 3 Airframe Selection

The main criteria for airframe selection are cost, availability, stability and maintenance. The airframe selected in the prototype development is radio controlled fixed wing trainer for aeromodelling hobby. This type of airframe is relatively low cost, easy to repair, and has a very good stability. Here are the specifications :

- wing span      : 1.8 meter
- weight   : 4.5 kg
- engine    : 9.9 cc / 2 horse power

The control axes are throttle (to control airspeed), aileron (to control roll rate or lateral axis), elevator (to control pitch rate or longitudinal axis) and rudder (to control yaw rate or directional axis).

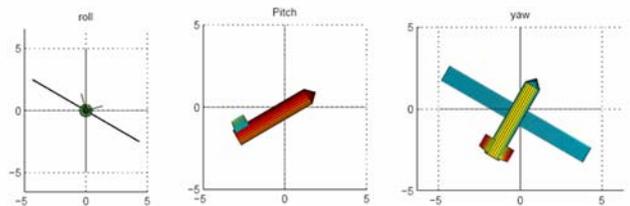

**Figure 4:** Roll, pitch and yaw

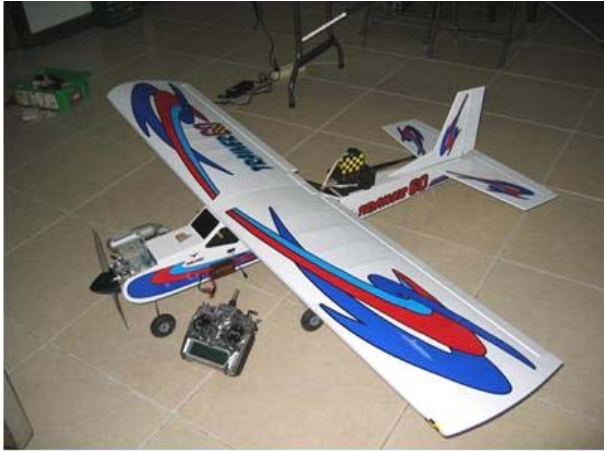

**Figure 5:** Airframe selected for UAV Test Bed

### 4 Control Algorithm

The control algorithm consists of two layers. The first upper layer is waypoint sequencer. The second lower layer is sets of PID (proportional, integrative, and derivative) controller. The waypoint sequencer reads the waypoints given to the autopilot control system by the operator. Each waypoint basically consists of 3D world coordinate which are latitude, longitude and altitude. Based on this waypoint information and current position, attitude and ground speed, the waypoint sequencer will output several objectives: attitude (roll, pitch and yaw/heading objective) and ground speed objectives. These objectives will be read by PID controller as its setting point and will be compared with actual value using PID algorithm to produce servo command value that will actuate the airframe's surface control (aileron, elevator and rudder) and throttle.

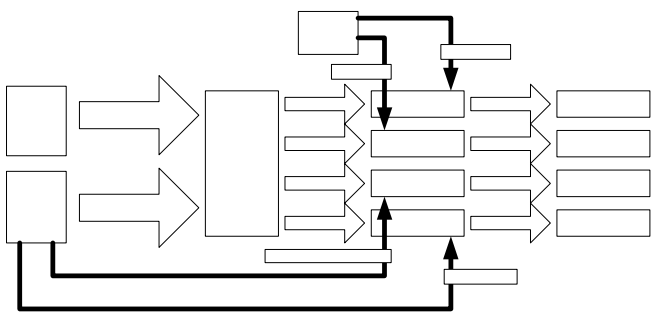

**Figure 6:** Block diagram of control algorithm

### 5 Sensor Selection

Based on the Control Algorithm Development step, there several measurements needed by the PID control scenarios. These measurements are position measurements and attitude measurements.

Position measurements are:

- speed,
- latitude,
- longitude, and
- altitude.

Attitude measurements are:

- heading / yaw,
- roll, and
- pitch.

For acquiring position measurements, a GPS receiver is used. The uBlox TIM-LA is chosen because it's relatively low cost and can provide 4 position information (speed, latitude, longitude, altitude and heading) every second (4Hz).

For measuring roll and pitch angle, the best solution would be using Attitude and Heading Reference System (AHRS). AHRS consists of inertial sensors (gyroscope and accelerometer) and magnetic field sensor (magnetometer). Strap down inertial navigation mechanization and proprietary fusion algorithm is usually used in combining the sensor readings to produce reliable attitude information. But the commercially available AHRS is beyond this research budget. The other simple and low cost alternative is using a pair of thermopile sensor for sensing the horizon. The idea comes from Co-Pilot$^{tm}$, an auxiliary device to train a beginner aeromodeller.

The basic principles are :

- thermopile sensor can sense remote object temperature,
- sky is typically having lower temperature than ground, and finally
- by installing the thermopile sensor in roll and pitch axis (4 thermopile sensors), during level flight all sensors approximately see the same amount of sky and ground, so the sensor output will approximately be the same. During pitch up (nose up) attitude the facing forward thermopile sensor sees more sky than ground, and the facing backward thermopile sensor sees more ground than sky. So the facing forward thermopile sensor sense cooler temperature than the facing backward thermopile sensor. By knowing the difference between the two sensor, the pitch up angle can be calculated. The same principle applied to the roll axis.

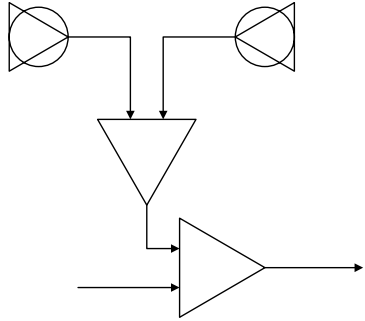

**Figure 7:** Thermopile Sensors Arrangement

The thermopile sensor used in this prototype is Melexis MLX90247. For sensing heading / yaw a rate gyroscope is used. The absolute heading offset for yaw rate integration is taken from GPS heading. The yaw rate gyro used in this prototype is Analog Device ADXRS401.

### 6 Autopilot Hardware Design and Prototyping

The UAV autopilot hardware system consists of two parts, the first part is for sensor processing and the second part is for stabilization and navigation control.

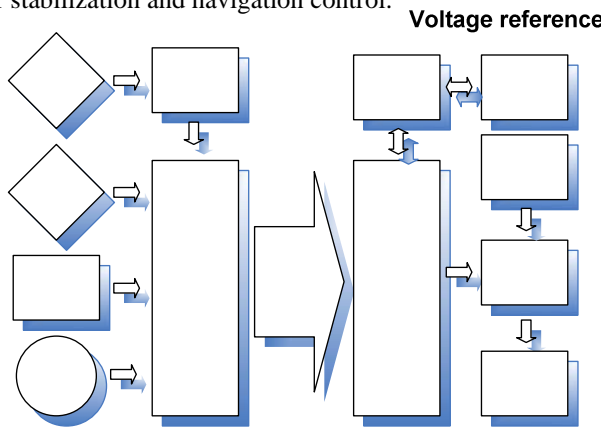

**Figure 8:** Autopilot hardware block diagram

### 7 Hardware in the Loop Simulation

Field trial is one of the most critical steps in UAV development. UAV usually consists of relatively high priced airframe, engine, actuator / servo, and payload system, so when there is failure in control system field trial, the risk is airframe crash, and usually only minor part of the crashed UAV that can be used for the next research and development. This step proved to be one of the main problems in UAV research and development.

One of the solutions for minimizing the effect of control system failure in field trial is Hardware in the Loop (HIL) Simulation.

### 7.1 HIL General Description

Hardware in the loop (HIL) simulator simulates a process such that the input and output signals show the same time-dependent values as the real dynamically operating components. This makes it possible to test the final embedded system under real working conditions, with different working loads and in critical/dangerous situations.

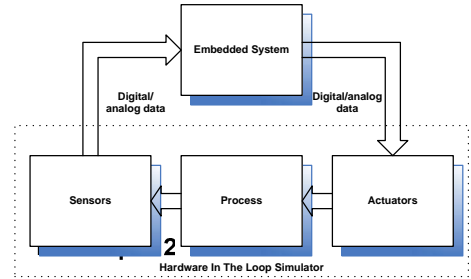

**Figure 9:** Autopilot hardware block diagram

In the case of UAV autopilot system development, A HIL simulator can be developed to simulate the flight characteristic of the airframe including the sensor output and the control input signal. The UAV autopilot system can be installed with the HIL simulator to see how the overall system works as a closed loop system. Here we can tune the PID gain parameter as well as the other system parameter and watch the effect to the airframe in the HIL simulator.

### 7.2 HIL Simulator Development

This writer develop HIL simulator based on commercially available simulation software. By using this approach, the basic simulation feature doesn't have to be implemented from scratch. Only specific functionality needed by HIL simulator need to be added. This specific functionality usually relates with interfacing between simulation software and autopilot hardware (sensor measurement and servo actuation simulation).

The chosen simulation software for HIL simulator development is X-Plane, because these reason:

- X-Plane is very interesting for non aerodynamicist developer, because we can make an airframe based only on its geometric dimension. The physics model is based on a process called Blade Element Theory. This set of principles breaks an airframe down by geometric shape and determines the number of stress points along its hull and airfoils. Factors such as drag coefficients are then calculated at each one of these areas to ensure the entire plane is being affected in some way by external forces. This system produces figures that are far more accurate than those achieved by taking averages of an entire airfoil, for example. It also results in extremely precise physical properties that can be computed very quickly during flight, ultimately resulting in a much more realistic flight model. The X-Plane accuracy of the flight model is already approved by FAA, for full motion simulator to train commercial airline pilot.

- X-Plane's functionality can be customized using a plug in. A plug in is executable code that runs inside X-Plane, extending what X-Plane does. Plug ins are modular, allowing developers to extend the simulator without having to have the source code to the simulator. Plug ins allow the extension of the flight simulator's capabilities or gain access to the simulator's data. For HIL simulator purposes we need to make plug in that

   a. reads attitude and position data to simulate sensor measurement

   b. writes surface control deflection values to simulate servo command

   c. has ability to communicate with autopilot hardware with some kind of hardware interface (to give sensor measurement and accept servo command).

### 7.3 X-Plane as HIL Simulator Platform
The HIL simulator consists of 3 main parts :

- sensors,
- process, and
- actuators.

The Sensors part simulates the sensor's output data in the airframe. This data will be processed by the UAV autopilot hardware as input. The sensor output data that should be produced by HIL simulator are position data (speed, altitude, latitude and longitude) and attitude data (roll, pitch and yaw). This can be accomplished by reading data from the simulator.

The Actuators part simulates how the UAV autopilot hardware can change the surface control of the airframe (aileron, elevator and rudder) and throttle position. In real world application this will be done by controlling the hobby servos put in the corresponding control surface or throttle engine. In HIL simulator this is done by writing data to the X-Plane that will affect the control surface of the simulated airframe.

The Process part simulates how the airframe will react to the input given by the UAV autopilot hardware.  So basically this part is where we should put the system dynamic model (transfer function). Generally this is the most complex part of the HIL simulator, but fortunately this part is already provided by the X-Plane (using its blade element approach). The HIL simulator plug in communicates with the UAV autopilot hardware through RS232 serial communication.

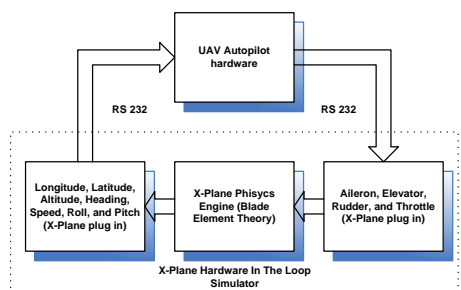

**Figure 10:    X-Plane as HIL simulation platform**

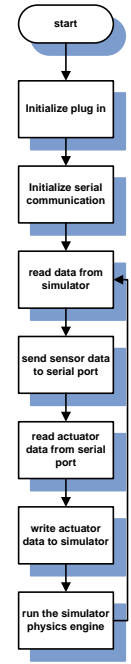

**Figure 11:    HIL simulator flow chart**

### 7.4 HIL Simulator Utilization
The HIL simulator is utilized to:

- develop the Ground Control Station Software,
- refine the firmware implementation through simulated closed loop tests,
- refine the hardware implementation through UAV autopilot long run reliability test, and
- PID gain tuning

There was one finding when testing the UAV autopilot reliability. The power supply regulator was not stable, and it can be seen from the overall system performance in the HIL simulator. This failure results in airframe crash in its worst. Since this test is conducted in HIL simulator no financial lost occurred. This is one example how HIL simulator can prevent airframe crash in real world field trial.

The HIL simulator enables the PID gain tuning based on trial end error basis. Analytical method of PID gain tuning is much more difficult since we have to have the mathematical model of the plant (airframe transfer function). It's considered easier for this writer to tune the PID gain  on trial and error basis since airframe crash is not a problem in HIL simulator.

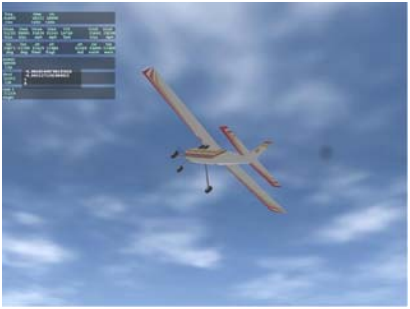

**Figure 12:** Hardware in the loop simulator 3D visualization

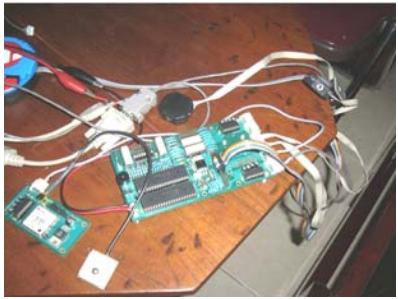

**Figure 13:** UAV autopilot hardware

Total time spent in the HIL simulation is comparable to 150 flight hours in real world field trial. After successful test in the HIL simulator the autopilot hardware is installed into UAV test bed prototype. This UAV prototype achieved its first successful autonomous flight in 19 January 2006, the mission is 6 waypoint autonomous navigation with loitering and crosstracking. Right now this autopilot system has been integrated into more than 6 kind of different airframes.

## 8 Concluding Remarks

The utilization of HIL simulator in the UAV development life cycle is proved to be very valuable. It enables the developer to test many aspects of the UAV autopilot hardware, finding the problem, refine the firmware, test the reliability, fine tune system parameter, and many others. All of those iterations are done inside development environment without risking the valuable airframe and payload. The HIL simulator also has the potential for low cost training tools for UAV operator.

## References

[1] Ronnback, Sven, *Development of a INS/GPS Navigation Loop for an UAV*, Institutionen for Systemteknik Avdelningen for Robotik och Automation, Lulea Tekniska Universitet, 2000

[2] Sanvido Marco,, *Hardware-in-the-loop Simulation Framework*, Automatic Control Laboratory, ETH Zurich

[3] Arya, Hemendra,, *Hardware-In-Loop Simulator for Mini Aerial Vehicle*, Centre for Aerospace Systems Design and Engineering, Department of Aerospace Engineering, IIT Bombay, India

[4] Gomez, Martin, *Hardware-in-the-Loop Simulation*, Embedded System Design, 2001, URL http://www.embedded.com/showArticle.jhtml?articleID=15201692

[5] Desbiens, Andre and Manai, Myriam,, *Identification of a UAV and Design of a Hardware-in-the-Loop System for Nonlinear Control Purposes*, Université Laval, Quebec City, Canada

[6] B. Taylor, C. Bil, and S. Watkins, 2003, *Horizon Sensing Attitude Stabilisation: A VMC Autopilot*, 18th International UAV Systems Conference, Bristol, UK,

[7] URL http://www.x-plane.com